\documentclass{bmvc2k}

\title{Learning Deep Representations of Appearance and Motion for Anomalous Event Detection}

\addauthor{Dan Xu$^1$}{danxuhk@gmail.com}{1}
\addauthor{Elisa Ricci$^{2,3}$}{eliricci@fbk.eu}{2} 
\addauthor{Yan Yan$^{1,4}$, Jingkuan Song$^1$}{yan@disi.unitn.it, jingkuan.song@unitn.it}{3}
\addauthor{Nicu Sebe$^1$}{sebe@disi.unitn.it}{4}

\addinstitution{
 $^1$DISI, University of Trento, \\ $\ \; $Trento, Italy}
\addinstitution{
$^2$Fondazione Bruno Kessler (FBK), \\$\ \; $Trento, Italy
}
\addinstitution{
$^3$University of Perugia, \\$\ \; $Perugia, Italy
}
\addinstitution{
$^4$ADSC, UIUC Singapore, \\$\ \; $Singapore
}

\runninghead{Xu et al.}{Learning Deep Representations of Appearance and Motion...}

\def\eg{\emph{e.g}\bmvaOneDot}

\def\etal{\emph{et al}\bmvaOneDot}

\def\eg{\emph{e.g}\bmvaOneDot}
\def\ie{\emph{i.e}\bmvaOneDot}

\def\etal{\emph{et al}\bmvaOneDot}
\newcommand{\RR}{I\!\!R}

\usepackage{multirow}
\usepackage{amssymb}
\usepackage{flushend}
\usepackage{epstopdf}
\usepackage{bm}

\begin{document}

\maketitle

\begin{abstract}
We present a novel unsupervised deep learning framework for anomalous event detection in complex video scenes. 
While most existing works merely use hand-crafted appearance and motion features, we propose Appearance and Motion 
DeepNet (AMDN) which utilizes deep neural networks to automatically learn feature representations. 
To exploit the complementary information of both appearance and motion patterns, 
we introduce a novel double fusion framework, combining both the benefits of traditional early fusion and late fusion strategies.
Specifically, 
stacked denoising autoencoders are proposed to separately learn both appearance and motion features as well as a joint representation 
(\textit{early fusion}). 
Based on the learned representations, multiple one-class SVM models are used to predict the anomaly scores of each input, 
which are then integrated with a 
\textit{late fusion} strategy for final anomaly detection. 
We evaluate the proposed method on two publicly available video surveillance datasets, showing competitive performance 
with respect to state of the art approaches.
\end{abstract}

\section{Introduction}
\label{sec:intro}
A fundamental challenge in intelligent video surveillance is to automatically detect abnormal events in long video streams. This problem
has attracted considerable attentions from both academia and industry in recent years 
\cite{lu2013abnormal,saligrama2012video,mahadevan2010anomaly,cong2sparse011}. Video anomaly detection is also important 
as it is related to other interesting topics in computer vision, such as 
dominant behavior detection~\cite{roshtkhari2013online}, visual saliency~\cite{zhai2006visual} and
interestingness prediction~\cite{grabner2013visual}.
A typical approach to tackle the anomaly detection task is to learn a model which describes normal activities in the video scene and then 
discovers unusual events by examining patterns which distinctly diverge from the model. However, the complexity of scenes and the 
deceptive nature of abnormal behaviours make anomaly detection still a very challenging task.
 
Among previous works, several anomaly detection approaches are based on analyzing individual moving objects in the scene. Tracking is usually an 
initial step for this class of methods. By using accurate tracking algorithms, trajectory extraction can be 
carried out to further perform trajectory clustering analysis~\cite{fu2005similarity, piciarelli2008trajectory} or design representative 
features \cite{stauffer2000learning} to model typical activities and subsequently discover anomalies. In \cite{wang2006learning},
trajectories which are spatially close and have similar motion patterns are identified and used for detecting unusual events.
Vaswani \etal~\cite{vaswani2005shape} propose a ``shape activity'' model 
to describe moving objects and detect anomalies. 
However, as tracking performance significantly degrades in the presence of several occluded targets, 
tracking-based methods are not suitable for analyzing complex and crowded scenes.  

To overcome the aforementioned limitations, researchers address the problem by learning spatio/temporal activity patterns 
in a local or a global context from either 2D image cells or 3D video volumes~\cite{cong2sparse011,mehran2009abnormal,benezeth2009abnormal,hospedales2012video,reddy2011improved,saligrama2012video, ricci2013prototype}. 
This category of methods builds the models based on hand-crafted features extracted from low-level appearance and motion cues, 
such as color, texture and optical flow. Commonly used low-level features include histogram of oriented gradients (HOG), 
3D spatio-temporal gradient, histogram of optical flow (HOF), among others. 
Cong \etal~\cite{cong2sparse011} employ multi-scale histograms of optical flow and a sparse coding model 
and use the reconstruction error as a metric for outlier detection.
Mehran \etal~\cite{mehran2009abnormal} propose a ``social force'' model based on optical flow features 
to represent crowd activity patterns and identify anomalous activities.
In~\cite{benezeth2009abnormal} co-occurrence statistics of spatio-temporal events are employed in combination with Markov 
Random Fields (MRFs) to discover unusual 
activities. Kratz \etal~\cite{kratz2009anomaly} introduce a HMMs-based approach for detecting abnormal events
by analyzing the motion variation of local space-time volumes.
Kim \etal~\cite{kim2009observe} propose a method based on local optical flow features and MRFs to spot unusual events. 
Mahadevan \etal~\cite{mahadevan2010anomaly} 
introduce a mixtures of dynamic textures model to jointly employ appearance and motion features. 
However, the adoption of hand-crafted features is a clear limitation of previous methods,
as it implies enforcing some some a priori knowledge which, in case of complex video
surveillance scene, is very difficult to define.

\begin{figure}[!t]
  \centering
  \includegraphics[width=4.82in]{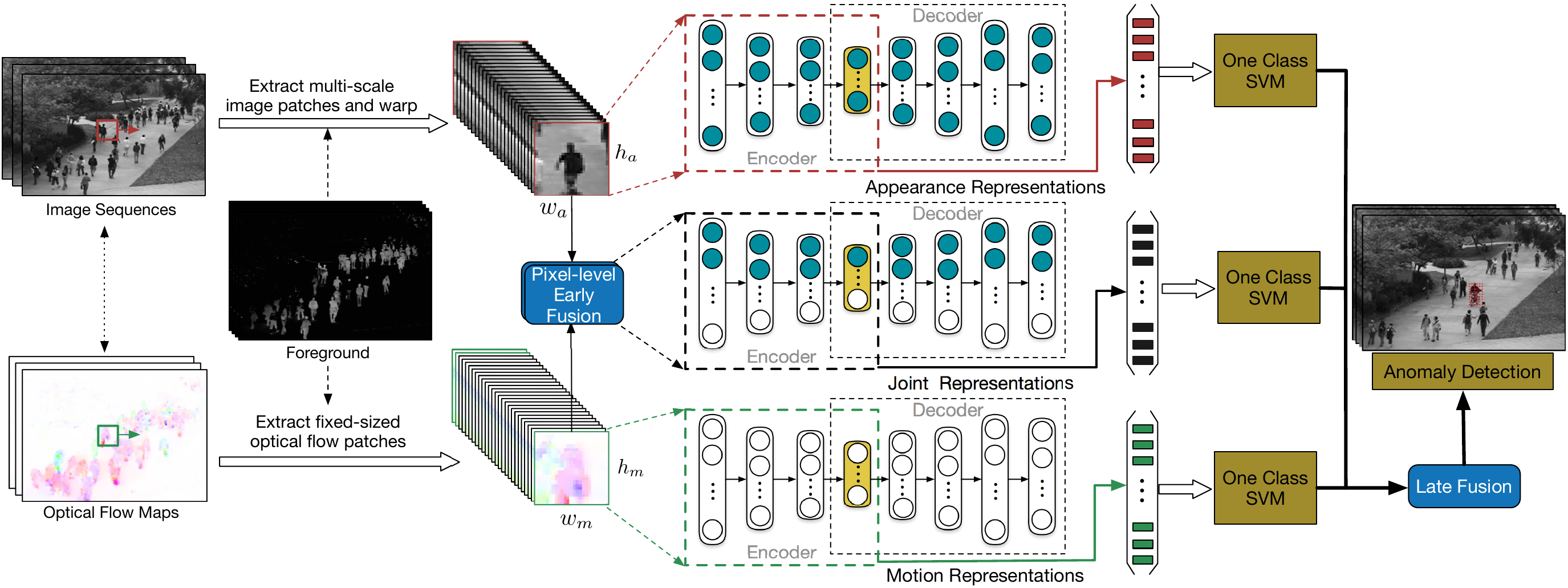}
  \caption{Overview of the proposed AMDN method for anomalous event detection.}
\label{fig:overview}
\vspace{-0.6cm}
\end{figure}

Recently, deep learning architectures have been successfully used to tackle various computer vision tasks, such as image 
classification~\cite{krizhevsky2012imagenet}, object detection \cite{girshick2014rich} and activity recognition~\cite{simonyan2014two}. 
However, these works are mainly based on Convolutional Neural Networks and consider a supervised learning scenario.
Unsupervised deep learning approaches based on autoencoder networks~\citep{vincent2008extracting} have also been investigated to 
address important tasks such as object tracking~\cite{wang2013learning} and face alignment~\cite{zhang2014coarse}. 
The key of the success is that, using deep architectures, rich and discriminative features can be learned via multi-layer 
nonlinear transformations. Therefore, it is reasonable to expect that detecting unusual events in videos can also benefit 
from deep learning models.

Following this intuition, in this paper we propose a novel Appearance and Motion DeepNet (AMDN) framework for discovering anomalous activities in complex video surveillance scenes.
Opposite to previous works~\cite{hospedales2012video,reddy2011improved,saligrama2012video}, instead of using hand-crafted features 
to model activity patterns, we propose to learn discriminative feature representations of both appearance and motion patterns
in a fully unsupervised manner. A novel approach based on stacked denoising autoencoders (SDAE)~\cite{vincent2010stacked} 
is introduced to achieve this goal. An overview of the proposed AMDN is shown in Fig.~\ref{fig:overview}. Low-level visual information 
including still image patches and dynamic motion fields represented with optical flow is used as input of two separate 
networks, to first learn appearance and motion features, respectively. To further investigate the correlations between appearance and motion, 
early fusion is performed by combining image pixels with their corresponding optical flow to learn a joint representation. 
Finally, for abnormal event prediction, a late fusion strategy 
is introduced to combine the anomaly scores predicted by multiple one-class SVM classifiers, each corresponding to one of the
three learned feature representations. The benefits of the proposed \textit{double fusion} framework (\ie combining both early fusion and late fusion
strategies) are confirmed by our extensive experimental evaluation, conducted on two publicly available datasets. 

In summary, the main contributions of this paper are: i) As far as we know, we are the first to introduce an unsupervised deep 
learning framework to automatically construct discriminative representations for video anomaly detection. ii) We propose a 
new approach to learn appearance and motion features as well as their correlations. Deep learning methods for combining 
multiple modalities have been investigated in previous works \cite{ngiam2011multimodal,srivastava2012multimodal}.
However, to our knowledge, this is the first work where multimodal deep learning is applied to anomalous event detection.
iii) A double fusion scheme is proposed to combine appearance and motion features for discovering unusual activities. It is worth noting 
that the advantages of combining early and late fusion
approaches have been investigated in previous works \cite{lan2012double}. However, Lan \etal.\cite{lan2012double} do not consider a deep 
learning framework.
iv) The proposed method is validated on 
challenging anomaly detection datasets and we obtain very competitive performance compared with the state-of-the-art.
\section{AMDN for Abnormal Event Detection}
The proposed AMDN framework for detecting anomalous activities is based on two main building blocks (Fig.\ref{fig:overview}).
First, SDAE are used to learn appearance and motion representations of visual data, as well as a joint representation capturing
the correlation between appearance and motion features (Sec. \ref{sec:autoencoder}). In the second phase (Sec. \ref{sec:anomaly}), to detect anomalous events, we propose to
train three separate one-class SVMs \cite{scholkopf2001estimating} based on the three different 
types of learned feature representations. Once the one-class SVM models are learned,
given a test sample corresponding to an image patch, three anomaly scores are computed and combined. The combination of the
one-class SVM scores is obtained with a novel late fusion scheme. In the following we describe the proposed approach in details.

\subsection{Learning Deep Appearance and Motion Representations}
\label{sec:autoencoder}
In this subsection we present the proposed AMDN for learning deep representations of appearance and motion. In the following we first introduce 
denoising autoencoders and then describe the details of the structure and the learning approach of the proposed AMDN.

\subsubsection{Denoising Autoencoders}
\label{sec:DAE}
A Denoising Auto-Encoder (DAE) is a one-hidden-layer neural network which is
trained to reconstruct a data point $\mathbf{x}_i$ from its (partially) corrupted version $\tilde{\mathbf{x}}_i$ \cite{vincent2008extracting}.
Typical corrupted inputs are obtained by drawing samples from a conditional distribution $p(\mathbf{x}|\tilde{\mathbf{x}})$ 
(\eg common choices for corrupting samples are additive Gaussian white noise or salt-pepper noise).
A DAE neural network can be divided into two parts: encoder and decoder, with a single shared hidden layer. These 
two parts actually attempt to learn two mapping functions, denoted as $f_e(\mathbf{W}, \mathbf{b})$ and 
$f_d(\mathbf{W'}, \mathbf{b'})$, where $\mathbf{W}, \mathbf{b}$ denote the weights and the bias term of the encoder part, 
and $\mathbf{W'}, \mathbf{b'}$ refer to the corresponding parameters of the decoder. For a corrupted input $\tilde{\mathbf{x}}_i$, a compressed 
hidden layer representation $\mathbf{h}_i$ can be obtained through 
$\mathbf{h}_i = f_e(\tilde{\mathbf{x}}_i\mid \mathbf{W}, \mathbf{b}) = \sigma(\mathbf{W}\tilde{\mathbf{x}}_i + \mathbf{b})$. Then, the 
decoder tries to recover the original input $\mathbf{x}_i$ from $\mathbf{h}_i$ computing 
$\hat{\mathbf{x}}_i = f_d(\mathbf{h}_i\mid \mathbf{W'}, \mathbf{b'}) = s(\mathbf{W'}\mathbf{h}_i + \mathbf{b'})$,
The function $\sigma(\cdot)$ and $s(\cdot)$ are activation functions, which are typically nonlinear transformations such as the sigmoid.
Using this encoder/decoder structure, the network can learn a more stable and robust feature representations of the input.

Given a training set $\mathcal{T}=\{\mathbf{x}_i\}_{i=1}^N$, a DAE learns its parameters $(\mathbf{W}, \mathbf{W'}, \mathbf{b}, \mathbf{b'})$ by solving the following regularized 
least square optimization problem:
\begin{equation}
\label{optimization_12}
\min_{\mathbf{W}, \mathbf{W}', \mathbf{b}, \mathbf{b}'} \ \ \sum_{i=1}^{N}\|\mathbf{x}_i-\mathbf{\hat{x}}_i\|_2^2 + 
\lambda (\|\mathbf{W}\|^2_F + \|\mathbf{W}'\|^2_F) 
\end{equation}
where $\|\cdot\|_F$ denotes the Frobenius norm. The first term represents the average reconstruction error,
while the weight penalty term is introduced for regularization. The parameter $\lambda$ balances the importance of the 
two terms. Typically, sparsity constraints are imposed on the output of the hidden units to discover meaningful representations 
from the data \cite{wang2013learning}. If we let $\mu_j$ be the target sparsity level and $\hat{\mu}_j = \frac{1}{N}\sum_{i=1}^{N} \mathbf{h}_i^j$ 
be the average activation values all over all training samples for the $j$-th unit, an extra penalty term based on cross-entropy,
$\varphi(\bm{\mu}||\hat{\bm{\mu}}) = -\sum^{H}_{j=1}[\mu_j \log(\hat{\mu}_j) + (1-\mu_j) \log(1-\hat{\mu}_j)]$, 
can be added to (\ref{optimization_12}) to learn a sparse representation. Here, $H$ is the number of hidden units.
The optimization problem (\ref{optimization_12}) has a non-convex objective function and gradient descent
can be used to compute a local optima. 

\subsubsection{AMDN Structure}
\begin{figure}[!t]
\centering
\includegraphics[width=4.4 in]{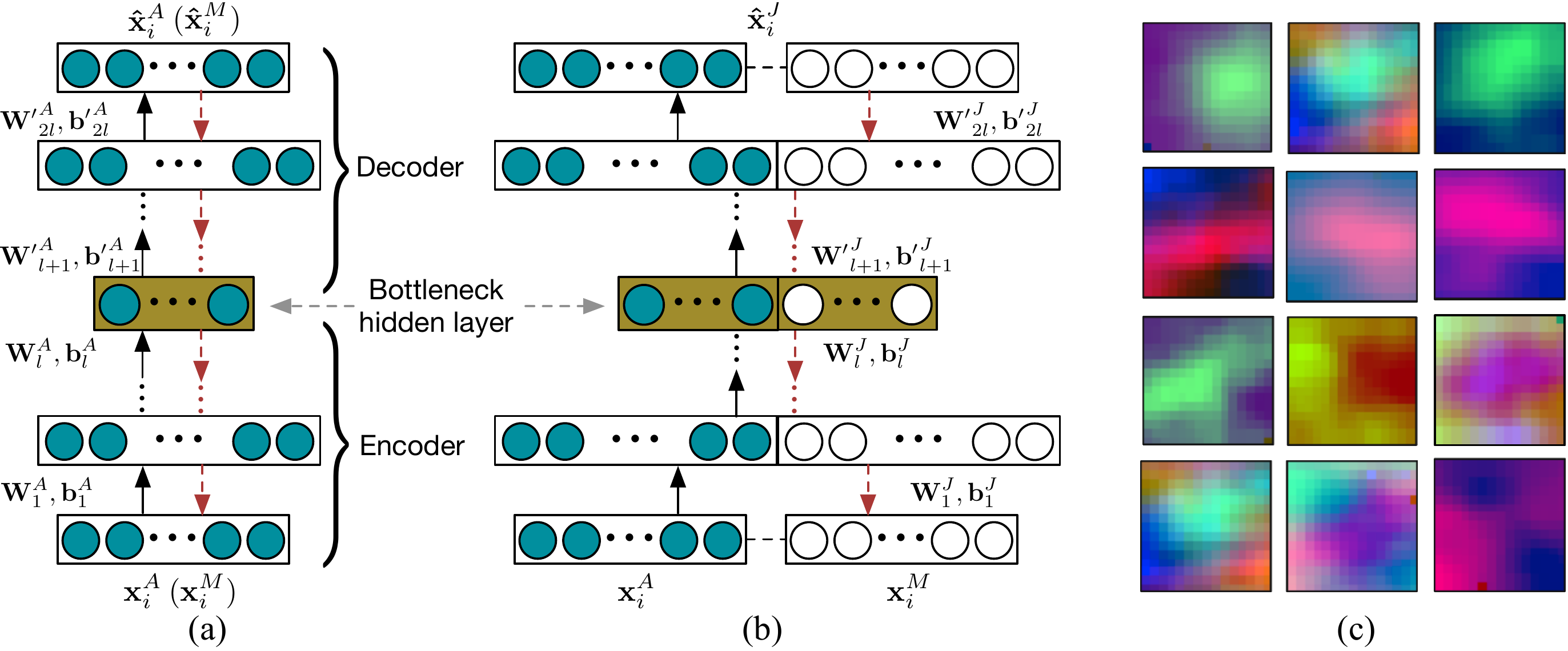} 
\vspace{-0.2cm}
\caption{(a) The structure of the appearance and motion representation learning pipelines. (b) The structure of the joint representation 
learning pipeline. (c) Some examples of the weight filters extracted from the first layer of the joint representation learning pipeline.}
\label{fig:filters}
\vspace{-0.6cm}
\end{figure}

The proposed AMDN structure consists of three SDAE pipelines (Fig.\ref{fig:overview}) corresponding to different types of low-level inputs. The three SDAE networks learn 
appearance and motion features as well as a joint representation of them. We show the basic structures of the proposed SDAE networks 
in Fig.~\ref{fig:filters} (a) and (b). Each SDAE consists of two parts: encoder and decoder. For the encoder part, we use an 
over-complete set of filters in the first layer to capture a representative information from the data. Then, the number of neurons 
is reduced by half in the next layer until reaching the ``bottleneck'' hidden layer. The decoder part has a symmetric structure with respect to 
the encoder part. 
We now describe the proposed three feature learning pipelines in details. 
\par
\textbf{Appearance representation.} This SDAE aims at learning mid-level appearance representations from the original image pixels. 
To capture rich appearance attributes, a multi-scale sliding-window approach with a stride $d$ is used to extract dense image patches, 
which are then warped into equal size $w_a \times h_a \times c_a$, where $w_a, h_a$ are the width and height of each patch and $c_a$ is the number of the channels ($c_a = 1$ for gray images). The warped 
patches $\mathbf{x}_i^{A} \in \RR^{w_a \times h_a \times c_a}$ are used for training. All the patches are linearly normalized into a range [0, 1]. We stack 4 encoding 
layers with $\nu_a \times w_a \times h_a \times c_a$ neurons in the first layer, where $\nu_a > 1$ is an amplification factor 
for constructing an over-complete set of filters. 
\par
\textbf{Motion representation.} The motion information is computed with optical flow. We use a sliding window approach with windows of 
fixed size $w_m \times h_m \times c_m$ ($c_m$ = 2 for optical flow magnitude along $x$ and $y$ axes), to generate dense optical 
flow patches $\mathbf{x}_i^{m} \in {\RR}^{w_m \times h_m \times c_m}$ for motion representation learning. Similar to the appearance feature
pipeline, the patches are normalized into [0,1] within each channel and 4 encoding layers are 
used. The number of neurons of the first layer is set to $\nu_m \times w_m \times h_m \times c_m$. 
\par
\textbf{Joint appearance and motion representation.} The above-mentioned SDAE learn appearance and motion features separately. 
Taking into account the correlations between motion and appearance, 
we propose to couple these two pipelines to learn a joint representation. The network training data $\mathbf{x}_i^{J} \in {\RR}^{w_j 
\times h_j \times (c_a + c_m)}$ are obtained through a 
pixel-level early fusion of the gray image patches and the corresponding optical flow patches. 

\subsubsection{AMDN Training}
We train the AMDN with two steps: pretraining and fine-tuning. The layer-wise pretraining learns one single denoising auto-encoder at a time using 
(\ref{optimization_12}) with sparsity constraints (Sec.\ref{sec:DAE}). The input is corrupted to learn the mapping function $f_e(\cdot)$, which 
is then used to produce the representation for the next layer with uncorrupted inputs. By using a greedy layer-wise pretraining, the 
denoising autoencoders can be stacked to build a multi-layer feedforward deep neural network, \ie a stacked denoising autoencoder. The network 
parameters are initialized through pretraining all layers, and then fine-tuning is used to adjust parameters over the whole network.

Fine-tuning treats all the layers of an SDAE as a single model. Given a training set 
$\mathcal{T}^{k} = \{\mathbf{x}_i^{k}\}_{i=1}^{N^{k}}$ with $N^{k}$ training samples ($k \in \{ A, M, J\}$ corresponds to 
appearance, motion and joint representation, respectively), the backpropagation algorithm can be used to fine-tune the network. 
The following objective function is used for fine-tuning the SDAE with $2L+1$ layers: 
\begin{equation}
J(\mathcal{T}^{k}) = \sum_i^{N^{k}} \|\mathbf{x}_i^k - \mathbf{\hat{x}}_i^k \|_2^2
+ \lambda_F \sum_{i=1}^{L} (\| \mathbf{W}_i^{k} \|_F^2 + \| \mathbf{W'}_i^{k} \|_F^2 ),
\end{equation}
where $\lambda_F$ is a user defined parameter. 
To speed up the convergence during training, stochastic 
gradient descent (SGD) is employed and the training set is divided 
into mini-batches with size $N_b^{k}$. 
Fig.~\ref{fig:filters}~(c) shows some of the learned filters in the first layer after fine-tuning for the joint representation learning pipeline.

After fine-tuning the whole network, the learned features 
representation can be computed to perform video anomaly detection. 
Theoretically, the output of each layer in an SDAE can be used as a novel learned feature representation. In this work, we 
choose the output of the ``bottleneck" hidden layer to obtain a more compact representation. Let $\mathbf{x}_i^{k}$ be the $i$-th input data sample, 
and $\sigma_l^{k}(\mathbf{W}_{l}^{k}, \mathbf{b}^{k}_{l})$ be the mapping function of the $l$-th hidden layer of the $k$-th SDAE pipeline. 
The learned features, $\mathbf{s}_i^{k}$, can be extracted through a forward pass computing 
$\mathbf{s}_i^{k}  = \sigma_{L}(\sigma_{L-1}(\cdots \sigma_1(\mathbf{W}_1^{k} \mathbf{x}^{k}_i + \mathbf{b}_1^{k})))$,
where the $L$-th hidden layer is the ``bottleneck" hidden layer.


\subsection{Abnormal Event Detection with Deep Representations}
\label{sec:anomaly}

We formulate the video anomaly detection problem as a patch-based binary categorization problem, \ie given a test 
frame we obtain $M_I \times N_I$ patches via sliding window with a stride $d$ and classify each patch as corresponding to a normal or abnormal region.
Specifically, given each test patch $t$ 
we compute three anomaly scores $A^{k}(\mathbf{s}^{k}_t)$, $k \in \{A, M, J\}$, using one-class SVM models and the
computed features representations $\mathbf{s}^{k}_t$.
The three scores are then linearly combined to obtain the final anomaly score $\mathcal{A}(\mathbf{s}^{k}_{t})=\sum_{k \in \{A, M, J\}} \alpha^{k}
A^{k}(\mathbf{s}^{k}_{t})$. A pre-processing of dynamic background subtraction can be carried out to improve the computing efficiency during the test phase as the anomalies exist in the foreground region.

\subsubsection{One-class SVM Modeling}
One-class SVM is a widely used algorithm for outlier detection, 
where the main idea is to learn a hypersphere in the feature space and map most of the training data into it. The outliers of the data
distribution correspond to point lying outside the hypersphere.
Formally, given a set of training samples $\mathcal{S}=\{\mathbf{s}_i^{k}\}_{i=1}^{N^k}$, the underlying problem of one-class 
SVM can be formulated as the following quadratic program: 
\begin{equation}\label{one_class_SVM}
\begin{split}	
    \min_{\mathbf{w}, \rho} & \ \ \ \frac{1}{2} \|\mathbf{w}\|^2 + \frac{1}{\nu N^k}\sum_{i=1}^{N^k}\xi_i - \rho
    \\ \textrm{s.t.} \ & \ \ \ \mathbf{w}^T\Phi(\mathbf{s}_i^{k}) \geq \rho - \xi_i, \ \xi_i \geq 0.
\end{split}
\end{equation} 
where $\mathbf{w}$ is the learned weight vector, $\rho$ is the offset, $\Phi(\cdot)$ is a feature projection function which maps 
feature vector $\mathbf{s}_i^{k}$ into a higher dimensional feature space. The user defined parameter $\nu \in (0, 1]$  
regulates the expected fraction of outliers distributed outside the hypersphere. 
Introducing a nonlinear mapping, the projection function $\Phi(\cdot)$ can be defined implicitely by introducing an associated kernel function
$k(\mathbf{s}_i^{k}, \mathbf{s}_j^{k})=\Phi(\mathbf{s}_i^{k})^T\Phi(\mathbf{s}_j^{k})$ and (\ref{one_class_SVM}) can be solved in the
corresponding dual form \cite{scholkopf2001estimating}. In our experiments we consider a rbf kernel, 
$k(\mathbf{s}_i^{k}, \mathbf{s}_j^{k}) =  e\frac{-\| \mathbf{s}_i^{k} -\mathbf{s}_j^{k} \| ^2}{2\sigma^2}$.
Given the optimal $\mathbf{w}$ and $\rho$ obtained by solving (\ref{one_class_SVM}), an outlier score for a test
sample $\mathbf{s}_t^{k}$ of the 
$k$-th SDAE pipeline can be estimated by computing $A^{k}(\mathbf{s}_t^{k}) = \rho - \mathbf{w}^T\Phi(\mathbf{s}_t^{k})$.

\subsubsection{Late Fusion for Anomaly Detection}
An unsupervised late fusion scheme is designed to automatically learn the weight vector
$\bm{\alpha}=[\alpha^{A}, \alpha^{M}, \alpha^{J}]$. The weight learning approach is based on the following optimization:
\begin{eqnarray}
\begin{array}{l}
\mathop {\min }\limits_{{\mathbf{W_s}^{k}}, \alpha^k}  \ \ \ \sum\limits_k {{\alpha^{k}}\mbox{tr}\left( {{\mathbf{W_s}^{k}}{\mathbf{S}^{k}}{{\left(
{{\mathbf{W_s}^{k}}{\mathbf{S}^{k}}} \right)}^T}} \right)}  + \lambda_s \left\| \bm{\alpha}  \right\|_2^2\\
\mbox{s.t.} \ \ \ \ \  {\alpha^{k}} > 0, \ \ \ \ \ \sum\limits_k {{\alpha^{k}}}  = 1
\end{array}
\label{eq.weight.obj}
\end{eqnarray}
where $\mathbf{S}^{k}=[\textbf{s}^k_1, \dots, \textbf{s}^k_{N^k}]$ is the matrix of training set samples, 
$\mathbf{W_s}^{k}$ maps the $k$-th feature $\textbf{s}^k_i$ into a new subspace and 
${{\mathbf{W_s}^{k}}{\mathbf{S}^{k}}\left( {{\mathbf{W_s}^{k}}{\mathbf{S}^{k}}} \right)} ^T$ represents the 
covariance of $k$-th feature type in the new subspace. The term $\left\| \bm{\alpha}  \right\|_2^2$ is introduced to avoid overfitting 
and $\lambda_s$ is a user defined parameter.
To solve (\ref{eq.weight.obj}), we first get $\mathbf{W_s}^{k}$ as the $d$ eigenvectors of $\mathbf{S}^{k}{\mathbf{S}^{k}}^T$
corresponding to the $d$-largest eigenvalues. Then, $\bm{\alpha}$ can be obtained by solving the simplex problem \cite{zoutendijk1960methods}:
\begin{equation}\label{opprox}
\mathop {\min }\limits_{{\alpha^{k}} > 0,\sum\limits_k {{\alpha^{k}}}  = 1} \frac{1}{2}{\left\| {\bm{\alpha} - \mathbf{c}} \right\|_2^2}
\end{equation}
with $\mathbf{c}=[c^{A}, c^{M}, c^{J}]$, ${c^{k}} =  - \frac{1}{{2\lambda_s }}\mbox{tr}\left( {{\mathbf{W_s}^{k}}{\mathbf{S}^{k}}
{{\left( {{\mathbf{W_s}^{k}}{\mathbf{S}^{k}}} \right)}^T}} \right)$.
Then for each patch $t$, we identify if it corresponds to an abnormal activity by computing the associated anomaly score
$\mathcal{A}(\mathbf{s}^{k}_t)$ and comparing it with a threshold $\eta$, \ie $\mathcal{A}(\mathbf{s}^{k}_t)\ \mathop{\lessgtr}\limits_{abnormal}^{normal} \ \eta$.

%
%
\vspace{-0.2cm}
\section{Experimental Results}
\begin{figure*}[!t]
  \centering
  \subfigure{\includegraphics[width=1.45in,height=2cm]{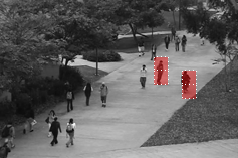}}
  \subfigure{\includegraphics[width=1.45in,height=2cm]{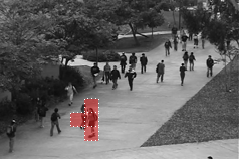}}
  \subfigure{\includegraphics[width=1.45in,height=2cm]{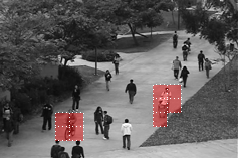}}\\\vspace{-3mm}
  \subfigure{\includegraphics[width=1.45in,height=2cm]{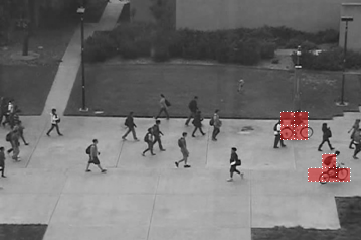}} 
    \subfigure{\includegraphics[width=1.45in,height=2cm]{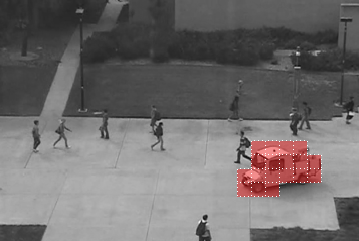}}
  \subfigure{\includegraphics[width=1.45in,height=2cm]{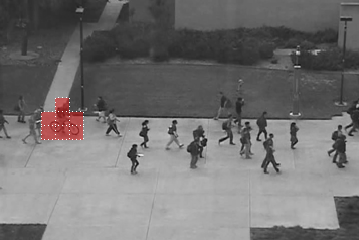}}
  \caption{Examples of anomaly detection results on Ped1 (top) and Ped2 (bottom) sequences. Our approach can detect 
  anomalies such as bikes, vehicles and skaters and provides accurate localization.} %
\vspace{-0.4cm}
  \label{examples}
\end{figure*}

\paragraph{Datasets and Experimental Setup.}
The proposed method is mainly implemented in Matlab and C++ based on Caffe framework \cite{jia2014caffe}. The code for optical flow calculation is written in C++ 
and wrapped with Matlab mex for computational efficiency \cite{liu2009beyond}. For one-class SVMs, 
we use the LIBSVM library (version 3.2)~\cite{chang2011libsvm}. The experiments are carried out on a PC with a middle-level graphics 
card (NVIDIA Quadro K4000) and a multi-core 2.1 GHz CPU with 32 GB memory. 
Two publicly available datasets, the UCSD (Ped1 and Ped2) dataset~\cite{mahadevan2010anomaly} and the Train dataset~\cite{zaharescu2010anomalous}, 
are used to evaluate the performance of the proposed approach.

The \textbf{UCSD pedestrian} dataset~\cite{mahadevan2010anomaly} is a challenging anomaly detection dataset including 
two subsets: Ped1 and Ped2. The video sequences depict different crowded scenes and anomalies include bicycles, 
vehicles, skateboarders and wheelchairs. In some frames the anomalies occur at multiple locations. 
Ped1 has 34 training and 16 test image sequences with about 3,400 anomalous and 5,500 normal frames, 
and the image resolution is $238\times158$ pixels. Ped2 has 16 training and 12 test image sequences
with about 1,652 anomalous and 346 normal frames. The image resolution is $360\times240$ pixels.

The \textbf{Train} dataset~\cite{zaharescu2010anomalous} depicts moving people inside a train. This is also a challenging abnormal event detection 
dataset due to dynamic illumination changes and camera shake problems. The dataset consists of 19218 frames, 
and the anomalous events are mainly due to unusual movements of people on the train.

\begin{figure*}[!t]
\centering
\subfigure[Frame-level ROC curve of PED1 Dataset]{\includegraphics[width=0.48\textwidth]{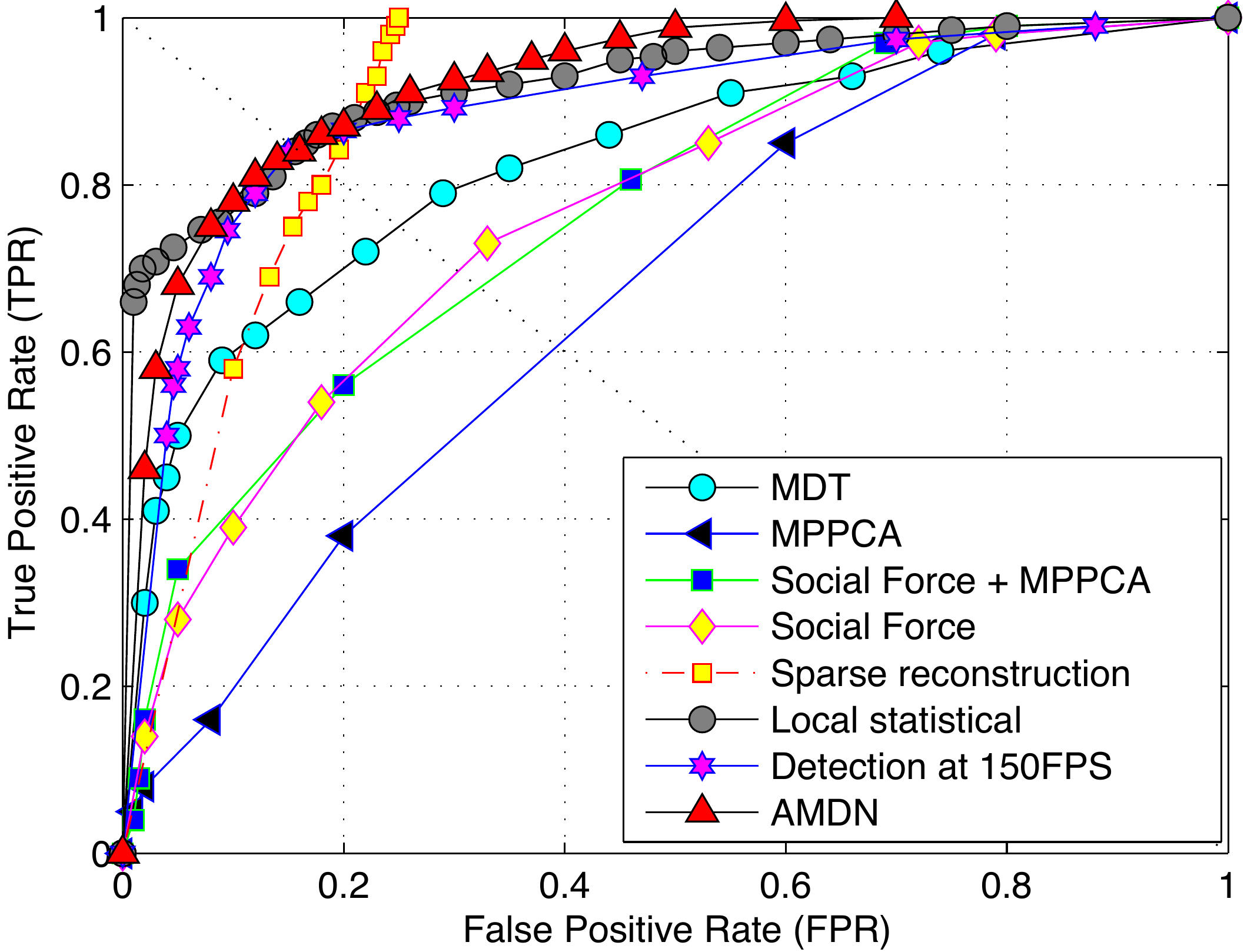}}
\subfigure[Pixel-level ROC curve of PED1 Dataset]{\includegraphics [width=0.48\linewidth]{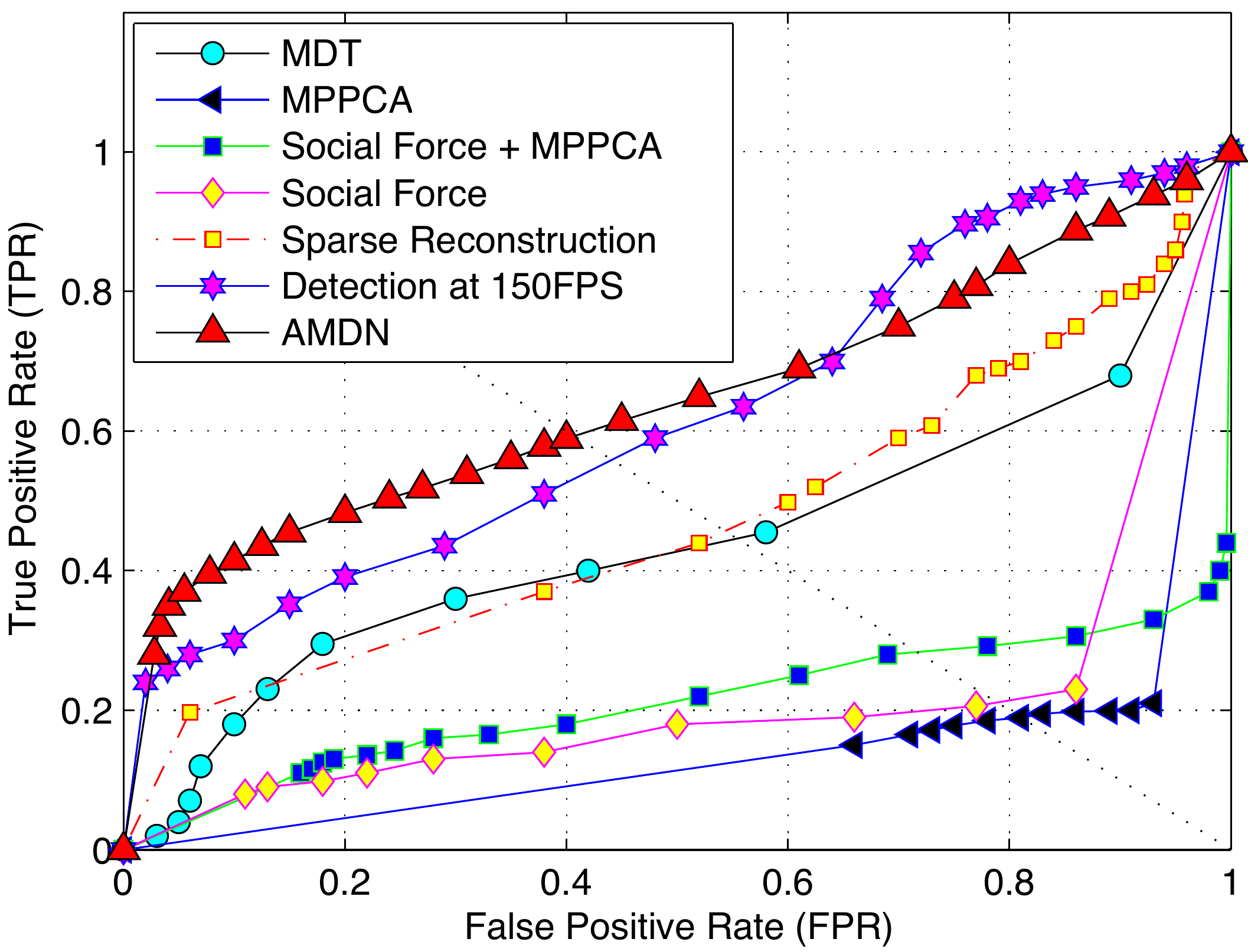}}
\caption{UCSD dataset (Ped1 sequence): comparison of frame-level and pixel-level anomaly detection results with state of the art methods. }
\label{ROC}
\vspace{-4pt}
\end{figure*}

\newcommand{\tabincell}[2]{\begin{tabular}{@{}#1@{}}#2\end{tabular}}
\begin{table}[!t]
\addtolength{\tabcolsep}{-1pt}  
\centering
\footnotesize
\resizebox{0.95\linewidth}{!} {
\begin{tabular}{lcccccc}
\hline
\multirow{2}{*}{Algorithm} & \multicolumn{2}{c}{Ped1(frame)} & \multicolumn{2}{c}{Ped1(pixel)} & \multicolumn{2}{c}{Ped2} \\ \cline{2-7}
& EER & AUC & EER & AUC & EER & AUC \\ \hline \hline
MPPCA \cite{kim2009observe} & 40\% & 59.0\% &81\%& 20.5\% & 30\% & 69.3\% \\
Social force \cite{mehran2009abnormal} & 31\% & 67.5\% & 79\%& 19.7\% & 42\% & 55.6\% \\
Social force+MPPCA \cite{mahadevan2010anomaly} & 32\% & 66.8\% &71\%& 21.3\% & 36\% & 61.3\%\\
Sparse reconstruction \cite{cong2sparse011} & 19\% & -- & 54\%& 45.3\% & -- & -- \\
Mixture dynamic texture \cite{mahadevan2010anomaly} & 25\% & 81.8\% & 58\%& 44.1\% & 25\% & 82.9\% \\
Local Statistical Aggregates \cite{saligrama2012video} & 16\% & 92.7\% &--& -- & -- & --\\
Detection at 150 FPS \cite{lu2013abnormal}& 15\% & 91.8\% & 43\%& 63.8\% & -- & --\\\hline\hline
\tabincell{c}{Joint representation (early fusion)} & \tabincell{c}{22\%} & \tabincell{c}{84.9\%} & 47.1\%& 57.8\%& \tabincell{c}{24\%} & \tabincell{c}{81.5\%}\\
\tabincell{c}{Fusion of appearance and motion pipelines (late fusion)} & \tabincell{c}{18\%} & \tabincell{c}{89.1\%} &43.6\%& 62.1\%& \tabincell{c}{19\%} & \tabincell{c}{87.3\%}\\
\tabincell{c}{AMDN (double fusion)} & \tabincell{c}{16\%} & \tabincell{c}{92.1\%} &40.1\%& 67.2\% & \tabincell{c}{17\%} & \tabincell{c}{90.8\%}  \\\hline
\end{tabular}
}
\vspace{0.1cm}
\caption{UCSD dataset: comparison in terms of EER (Equal Error Rate) and AUC (Area Under ROC) 
with the state of the art methods on Ped1 and Ped2.}
\label{AUC}
\vspace{-0.36cm}
\end{table}
\vspace{-0.52cm}

\paragraph{Quantitative evaluation.}
In the first series of experiments we evaluate the performance of the proposed method on the UCSD dataset.
For appearance learning, patches are extracted using a sliding window approach at three different scales, \ie $15 \times 15$, $18\times 18$ 
and $20\times 20$ pixels. This generates more than 50 million image patches, 10 million of which are randomly sampled and 
warped into the same size ($w_a \times h_a = 15 \times 15$ pixels) for training. For learning the motion representation, 
the patch size is fixed to $w_m \times h_m = 15 \times 15$ pixels, and 6 million training patches are randomly sampled. In the test phase, we use sliding widow with a size of $15 \times 15$ and a stride $d = 15$. The number of neurons of the first layer 
of the appearance and motion network is both set to 1024, while for the joint pipeline is 2048. Then the encoder part 
can be simply defined as: $1024 (2048) \rightarrow 512 (1024) \rightarrow 256 (512) \rightarrow 128(256)$, and the decoder part is 
a symmetric structure. For the pre-training of the DAE, the corrupted inputs are produced by adding a Gaussian noise with variance 0.0003. 
The network training is based on the SGD with a momentum parameter set to 0.9. We use a fixed learning rates $\lambda=0.01$, $\lambda_F = 0.0001$ 
and a mini-batch size $N_b = 256$. For one-class SVMs, the parameter 
$\nu$ is tuned with cross validation. The learned late fusion weights $[\alpha^A, \alpha^M, \alpha^J]$ are obtained with $\lambda_s=0.1$ and for Ped1 and Ped2 are $[0.2, 0.5, 0.3]$ 
and $[0.2, 0.4, 0.4]$, respectively.

Some examples of anomaly detection results on the UCSD dataset are shown in Fig.~\ref{examples}. To perform a quantitative evaluation, 
we use both a frame-level ground truth and a pixel-level ground truth. The frame-level ground truth represents whether one or 
more anomalies occur in a test frame. The pixel-level ground truth is used to evaluate the anomaly localization performance. 
If the detected anomaly region is more than 40\% overlapping with the annotated region, it is a true detection. 
We carry out a frame-level evaluation on both Ped1 and Ped2. Ped1 provides 10 test image sequences with pixel-level ground truth. 
The pixel-level evaluation is performed on these test sequences.

Fig.~\ref{ROC} (a) and (b) show the frame-level and pixel-level detection results on Ped1. The ROC curve is produced 
by varying the threshold parameter $\eta$. Table~\ref{AUC} shows a quantitative comparison in terms of 
Area Under Curve (AUC) and Equal Error Rate (EER) of our method with several state-of-the-art approaches. From the frame-level evaluation, 
it is evident that our method outperforms most previous methods and that its performance are competitive with the best two baselines  
\cite{saligrama2012video} and~\cite{lu2013abnormal}. 
Moreover, considering pixel-level evaluation, \ie accuracy in anomaly localization, our method outperforms all the competing approaches 
both in terms of EER and AUC. 
Table~\ref{AUC} also demonstrate the advantages of the proposed double fusion strategy.
Our AMDN guarantees better performance than early fusion and late fusion approaches. Specifically,
for early fusion we only consider the learned joint appearance/motion representation 
and a single one-class SVM. For late fusion we use the two separate appearance and motion pipelines and the proposed fusion scheme but we discard
the joint representation pipeline. Interestingly, in this application the late fusion strategy outperforms early fusion. 

\begin{figure}[!t]
\centering
\includegraphics[width=4.8in]{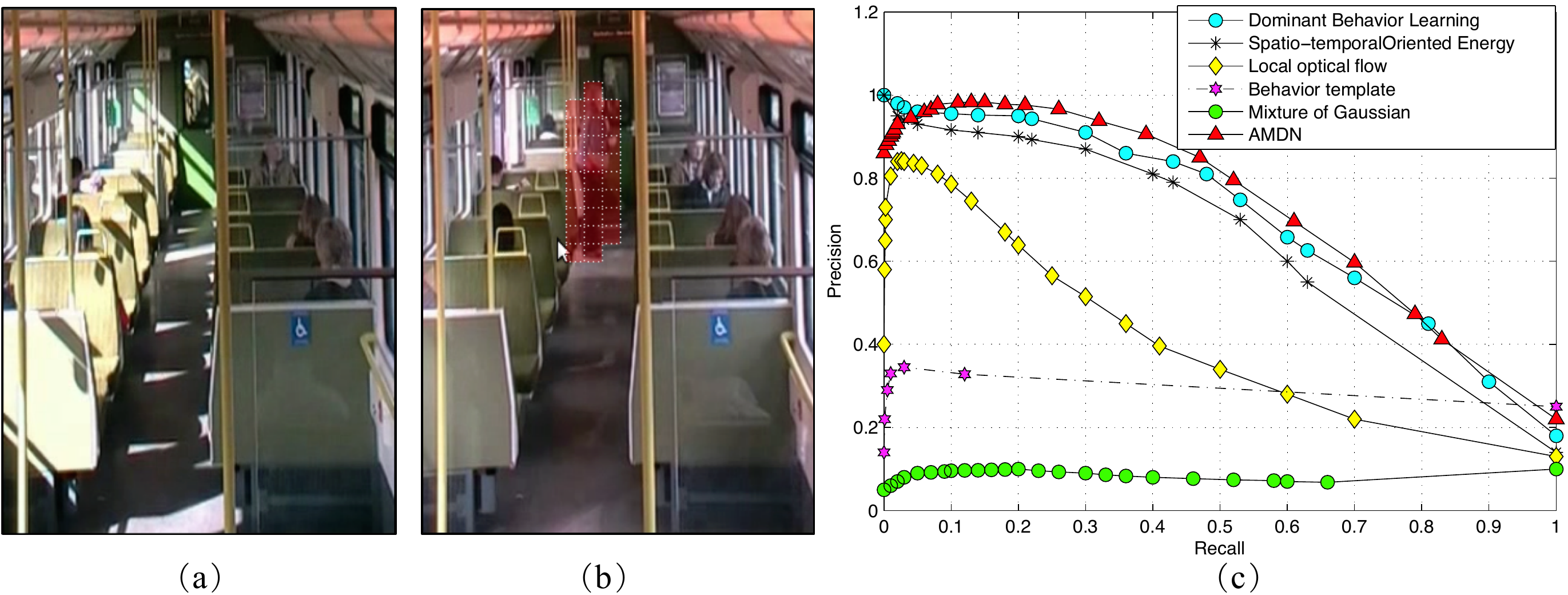} 
\caption{Train dataset: (a) a frame depicting typical activities, (b) an example of a detected anomaly. 
(c) precision/recall curve.}
\label{traindataset}
\vspace{-0.3cm}
\end{figure}

In the second series of experiments we consider the Train dataset. For parameters, we use the same experimental setting
of the UCSD experiments, except from the parameters $\lambda_F$ and $N_b$ which are set to $0.00001$ and $100$, respectively. 
The learned late fusion weights are $[\alpha^A, \alpha^M, \alpha^J]=[0.3, 0.4, 0.3]$. We compare the proposed approach with several methods, 
including dominant behavior learning~\cite{roshtkhari2013online}, spatio-temporal oriented energies~\cite{zaharescu2010anomalous}, local optical flow~\cite{adam2008robust}, 
behavior templates~\cite{jodoin2008modeling} and mixture of Gaussian. From the precision/recall curve shown in Fig.~\ref{traindataset} (c), 
it is clear that our method outperforms all the baselines. 


%

%
\section{Conclusions}
We presented a novel unsupervised learning approach for video anomaly detection based on
deep representations. The proposed method is based on multiple stacked autoencoder networks for learning both appearance and motion representations of scene activities. A double
fusion scheme is designed to combine the learned feature representations. Extensive experiments on two challenging 
datasets demonstrate the effectiveness of the proposed approach and show competitive performance with respect to existing methods. Future works 
include investigating other network architectures, alternative approaches for fusing multimodal data in the context of SDAE, and extending our framework using multi-task learning~\cite{yan2013no,yanMulti-task} for detecting anomalies in heterogeneous scenes.

\section*{Acknowledgments}
This work was partially supported by the MIUR Cluster project Active Ageing at Home, the EC H2020 project ACANTO and by A*STAR Singapore under the Human-Centered Cyber-physical Systems (HCCS) grant. The authors also would like to thank NVIDIA for GPU donation.

\bibliography{bmvc}
\end{document}